\documentclass{article}

\usepackage{microtype}
\usepackage{graphicx}
\usepackage{subfigure}
\usepackage{booktabs} 

\usepackage{hyperref}

\usepackage[accepted]{icml2025}

\usepackage{amsmath}
\usepackage{amssymb}
\usepackage{mathtools}
\usepackage{amsthm}
\usepackage{multicol}
\usepackage{multirow}

\usepackage[capitalize,noabbrev]{cleveref}

\theoremstyle{plain}

\theoremstyle{definition}

\theoremstyle{remark}

\newcommand{\argmin}{\mathop{\mathrm{argmin}}}

\usepackage[textsize=tiny]{todonotes}

\icmltitlerunning{Data-Incremental Continual Offline Reinforcement Learning}

\begin{document}

\twocolumn[
\icmltitle{Data-Incremental Continual Offline Reinforcement Learning}



\icmlsetsymbol{equal}{*}

\begin{icmlauthorlist}
\icmlauthor{Sibo Gai}{yyy}
\icmlauthor{Donglin Wang}{yyy}
\end{icmlauthorlist}

\icmlaffiliation{yyy}{School of Engineering, Westlake University, Hangzhou, China}

\icmlcorrespondingauthor{Sibo Gai}{gaisibo@westlake.edu.cn}

\icmlkeywords{continual learning, reinforcement learning, continual reinforcement learning}

\vskip 0.3in
]



\printAffiliationsAndNotice{}  

\begin{abstract}
In this work, we propose a new setting of continual learning: data-incremental continual offline reinforcement learning (DICORL), in which an agent is asked to learn a sequence of datasets of a single offline reinforcement learning (RL) task continually, instead of learning a sequence of offline RL tasks with respective datasets. Then, we propose that this new setting will introduce a unique challenge to continual learning: active forgetting, which means that the agent will forget the learnt skill actively. The main reason for active forgetting is conservative learning used by offline RL, which is used to solve the overestimation problem. With conservative learning, the offline RL method will suppress the value of all actions, learnt or not, without selection, unless it is in the just learning dataset. Therefore, inferior data may overlay premium data because of the learning sequence. To solve this problem, we propose a new algorithm, called experience-replay-based ensemble implicit Q-learning (EREIQL), which introduces multiple value networks to reduce the initial value and avoid using conservative learning, and the experience replay to relieve catastrophic forgetting. Our experiments show that EREIQL relieves active forgetting in DICORL and performs well.
\end{abstract}

\section{Introduction}
Today, offline reinforcement learning (RL) is an area of interest. Like a human being, with the offline RL, a robot can learn from datasets precollected by other robots or human beings without exploring the real world by itself. In this way, the robot can learn safety and efficiency. As a data-based machine learning method, the performance of the offline RL is highly affected by the richness and quality of the data. To update and stay optimal throughout its whole lifetime, an agent must have the ability to leverage and learn every data it encounters. For example, a dash-washer robot is watching a video of a kitchen. By observing different kitchen workers during their lifetimes, the robot can optimise the origin policy (such as cleaning well) and broaden the scope of known skills (such as cleaning new types of cookingware). All these skills belong to a single task: dash cleaning, but related to different datasets that are collected at different times. In response to this demand, we propose that the offline RL should have a new capability: the lifelong evolution ability of a single task.

\begin{figure}[h]
	\centering
	\includegraphics[width=\linewidth]{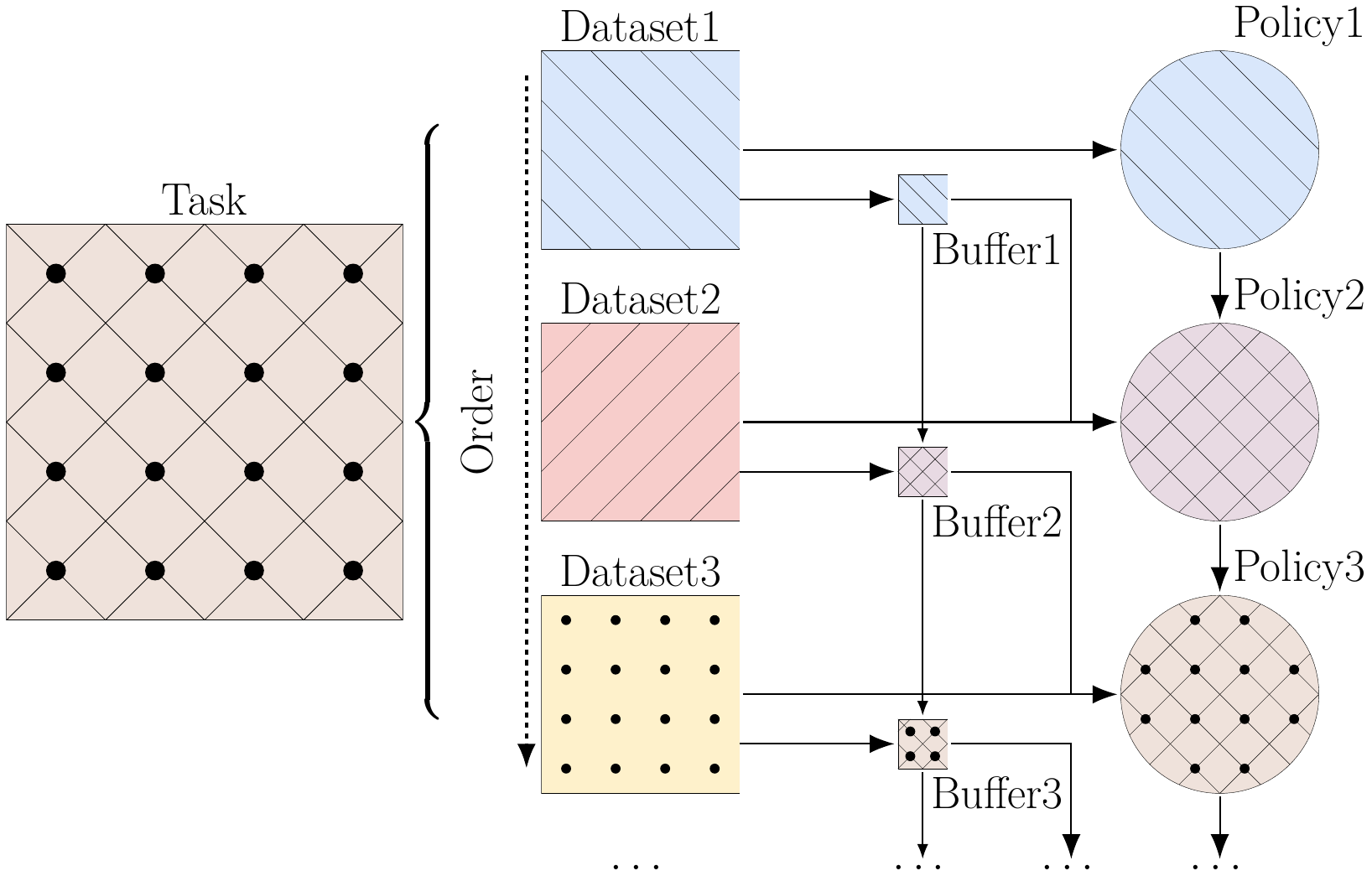}
	\caption{The diagram of DICORL. The algorithm needs to learn a sequence of datasets of a single task sequentially and expect to perform best on the task as a whole, rather than on individual datasets.}
	\label{fig:STCORL_single_task}
\end{figure}

In this work, we propose a new setting that combines continual learning and offline RL: data-incremental continual offline reinforcement learning (DICORL). In the DICORL setting, the agent should learn a sequence of datasets of a single RL task continually and get an optimal result in the task in all of its lifetime. In comparison, we call the traditional continual offline RL \cite{gaiOfflineExperienceReplay2023,huang2024solvingcontinualofflinereinforcement} which learns a sequence of offline RL tasks continually the task-incremental continual offline reinforcement learning (TICORL). We summarise the DICORL setting in Fig.~\ref{fig:STCORL_single_task}. As shown in the figure, there is a sequence of datasets belonging to a single task, each of them representing part of the characteristic of the task. These datasets may have different sources and, therefore, have different state spaces and behaviour policies. The agent should learn from these datasets sequentially. However, instead of learning a policy that can reflect each dataset, the purpose of DICORL is to learn the task itself. This is the main characteristic and challenge of DICORL. Compared with traditional TICORL, the DICORL algorithm needs not only to remember all the past skills but also to synthesise them. This is also the reason why data-incremental continual learning is a unique problem for offline RL: An image classification task or an online RL task usually focusses on quality-homogeneous datasets, which will not meet the problem of synthesis.

However, in this work, we propose that there is a new problem that the DICORL algorithm must overcome: active forgetting. Active forgetting is a unique problem that only occurs in DICORL, caused by a major technology used by offline RL: conservative learning. Conservative learning is the most dominant method to solve the overestimation problem, which is the major problem of offline RL. Specifically, conservative learning methods like \cite{kumarConservativeQLearningOffline2020} will penalise the value of actions not in the datasets. In DICORL, because the agent cannot distinguish the fresh-new out-of-distribution (OOD) data (which does not belong to any dataset) and those data that have been learnt in one or some of the past datasets, the conservative learning method will actively penalise all the OOD data as well as the learned one. In this way, as a result, the agent can only learn the final dataset without memory. If the high-performance dataset is learnt first, the final performance will be affected. Another important category of offline RL methods, the policy constraint methods like TD3+BC\cite{fujimotoMinimalistApproachOffline2021}, will also face the active forgetting in a much straight way: they will clone the data from new tasks without comparison.


Because active forgetting is not a simple distribution shift problem, simply introducing the traditional continual learning method into DICORL cannot solve the problem. Therefore, we propose a new algorithm based on \cite{kostrikovOfflineReinforcementLearning2021a} and \cite{anUncertaintyBasedOfflineReinforcement2021}, called experience-replay-based ensemble implicit Q-learning (EREIQL). Specifically, EREIQL introduces an ensemble value function. By initialising multiple value functions, EREIQL assigns a sufficiently low value initialisation value to each state, so that EREIQL can overcome the overestimation problem without conservative learning. Meanwhile, the policy in EREIQL adopts advantage weighted regression with experience replay to avoid learning inferior actions with lower Q-values and relieve catastrophic forgetting.

The main contributions of this work are: 1) we propose a new setting DICORL and point out that existing offline RL algorithms will lead to active forgetting in the DICORL problem; 2) we put forward a new EREIQL algorithm that avoids active forgetting through passively conservative learning; 3) experiments on various datasets show that EREIQL proposed here can achieve superior performance on different DICORL tasks.

\section{Related Work}
\subsection{Continual Learning}
Continual learning aims to use a single network to continuously learn multiple tasks while consuming acceptable resources, enabling excellent performance across tasks. These algorithms can be broadly categorized into three classes: rehearsal-based, regularization-based, and dynamic-architecture-based. Our work belongs to rehearsal-based methods, so we will introduce it specifically.


Rehearsal-based continual learning maintains performance on previous tasks by retaining some data from them in a replay buffer and using it when learning new tasks. The first critical issue here is the construction of the replay buffer. Related research includes randomly storing \cite{bangRainbowMemoryContinual2021,prabhuGDumbSimpleApproach2020} data from different tasks and selective storing \cite{yoonOnlineCoresetSelection2021,hayesSelectiveReplayEnhances2021} based on characteristics such as value, uniqueness, and representativeness. Another focal research direction is to use selected data, most often blending it with new task data in new batches to learn \cite{rolnickExperienceReplayContinual2019} or distilling knowledge from old data and old networks retained \cite{smithAlwaysBeDreaming2021}. Another line of methods uses a generator to produce samples following the same distribution as the task data instead of directly storing a replay buffer \cite{luoLearningPredictGradients2022,millichampBraininspiredFeatureExaggeration2021}. These approaches avoid occupying space with old task data, but increase overall complexity.
\subsection{Continual Reinforcement Learning}
Continual RL methods mainly focus on two aspects.
First, how to select data for storage. Related work here includes \cite{hayesSelectiveReplayEnhances2021,kaplanisContinualReinforcementLearning2020}, seeking to retain the most critical data points, using strategies such as choosing the highest value experiences and averaging sampling in state space where possible.
The second focal issue in CRL is integrating continual learning into RL algorithms. The main work here includes \cite{wolczykContinualWorldRobotic2021,wolczykDisentanglingTransferContinual2022}, showing that continual learning can play a role in RL, working better in the actor network than in the critic network. \cite{chenForgettingLessGood2023} notes that although continual learning aims to alleviate forgetting, existing methods that reduce forgetting can simultaneously enhance forward transfer capabilities. For more details, refer to the survey \cite{lesortContinualLearningRobotics2020} and the foundational work \cite{ringFormalFrameworkContinual2005,abelDefinitionContinualReinforcement2023}.

\subsection{Offline Reinforcement Learning}
Offline RL refers to the RL approach in which agents learn skills not through interacting with the environment, but from offline datasets of experiences and trajectories collected from other agents or humans. The most critical problem that agents need to solve in this learning method is overestimation.
Initially, proposed solutions to this problem constrain the deviation between the learnt policy and that in the offline data. These algorithms include \cite{pengAdvantageWeightedRegressionSimple2019,nairAWACAcceleratingOnline2021,zhuangBehaviorProximalPolicy2023}, which incorporate a KL divergence constraint during policy learning, limiting the discrepancy between agent policy and offline policy. \cite{kumarStabilizingOffPolicyQLearning2019} suggests this deviation should emphasise actions whose behaviour policy probability is zero, requiring the probability of selecting these actions also to be zero. Furthermore, \cite{fujimotoMinimalistApproachOffline2021} shows that this deviation can be achieved simply by appending a behaviour cloning term to online RL algorithms.

However, since these methods restrict the distance between the learnt and behaviour policies, they are susceptible to offline data quality \cite{maConservativeOfflineDistributional2021}. Another line of RL algorithms addresses this issue by learning a conservative Q function. These include conservative Q-learning (CQL) proposed by \cite{kumarConservativeQLearningOffline2020}. CQL avoids direct policy constraints, addressing the data quality problem by enforcing lower values for OOD data. Other approaches like \cite{kostrikovOfflineReinforcementLearning2021a} concurrently leverage value and Q functions, averting overestimation by using only Q values in-distribution from the samples. Instead of reducing the Q values of the OOD data through constraints, \cite{anUncertaintyBasedOfflineReinforcement2021} assigns lower initial values through ensemble learning for conservative learning.

Combine offline RL and continual RL, these days some work is already focused on continual offline RL, including \cite{gaiOfflineExperienceReplay2023,huang2024solvingcontinualofflinereinforcement}. All of these works are focused on TICORL, but not on our DICORL setting.
\section{Problem Definition and Preliminary}
\subsection{Single Task Continual Offline Reinforcement Learning}
Unlike traditional continual learning, DICORL learns only one offline RL task $T$. The offline RL task can be formulated as a Markov decision process (MDP) tuple $\{\mathcal{S},\mathcal{A},P,\rho_0,r,\gamma\}$, where $\mathcal{S}$ denotes the state space, $\mathcal{A}$ the action space, $P:\mathcal{S}\times\mathcal{A}\times\mathcal{A}\rightarrow\left[0,1\right]$ the transition probability, $\rho_0:\mathcal{S}$ the initial state distribution, $r:\mathcal{S}\times\mathcal{A}\rightarrow\left[-R_\mathrm{max},R_\mathrm{max}\right]$ the reward function, and $\gamma\in\left(0,1\right]$ the discount factor. The final return is the cumulative discount reward during the motion $R_{t,n}=\sum_{i=t}^H\gamma^{\left(i-t\right)}r\left(\mathbf{s}_i,\mathbf{a}_i\right)$ with $H$ denoting the maximum execution steps.

We employ an actor-critic RL architecture which achieves the best performance in offline RL. The actor-critic RL consists of three parameterized networks: a critic network $Q\left(\mathbf{s},\mathbf{a}\right)$, a value function $V\left(\mathbf{s}\right)$, and an actor network $\pi\left(\mathbf{a}|\mathbf{s}\right)$. Q learning trains the critic network using the Bellman operator: $\mathcal{B}^*Q\left(\mathbf{s},\mathbf{a}\right)=r\left(\mathbf{s},\mathbf{a}\right)+\gamma\mathbb{E}_{\mathbf{s}^\prime\sim P\left(\mathbf{s}^\prime|\mathbf{s},\mathbf{a}\right)}V\left(\mathbf{s}^\prime\right)$ and $V\left(\mathbf{s}^\prime\right) = \max_{\mathbf{a}^\prime}Q\left(\mathbf{s}^\prime,\mathbf{a}^\prime\right)$.

In offline RL, an agent learns not through environmental interaction but from offline datasets. In DICORL, the agent needs to learn from a series of offline datasets $\mathcal{D}=\left\{D_1,\dots,D_N\right\}$. Each dataset $D_n=\left(\mathbf{s}_{i,n},\mathbf{a}_{i,n},\mathbf{s}_{i,n}^\prime,r_{i,n}\right), i\in I$ is commonly assumed to be sampled from some (unknown) behaviour policy $\pi_n^\beta$, $I$ is the index set. These behaviour policies can be seen as independently and identically distributed (i.i.d.) samples from a distribution $\mathcal{\Pi}^\beta$, with no information on relationships among behaviour policies available to the algorithm. For convenience, subscripts $n$ denoting specific tasks are omitted in the remainder when a particular task does not need to be specified.

\subsection{Conservative Learning and Active Forgetting}

RL uses the Bellman equation to update the Q function. As a bootstrap method, Q learning uses a maximum future value to optimise itself. In offline RL, if the value of the future does not belong to the dataset, it will never be optimised. Once this OOD state prevails over other states in the value mistakenly, it will accumulate to the Q values of all preceding states through bootstrapping and cause the overestimation problem.

To address this problem, a successful class of offline RL algorithms such as Conservative Q-Learning (CQL) \cite{kumarConservativeQLearningOffline2020} employs conservative learning. Specifically, these algorithms actively suppress the Q values of actions selected by the policy during learning while boosting the Q values for actions present in the dataset:  

\begin{eqnarray}
Q^\text{update}&=&\argmin\limits_{Q}\alpha_\text{CQL}\left(\mathbb{E}_{\mathbf{s}\in D,\mathbf{a}\sim\pi\left(\mathbf{a}|\mathbf{s}\right)}\left[Q\left(\mathbf{s},\mathbf{a}\right)\right]\right.\nonumber\\&&-\left.\mathbb{E}_{\mathbf{s}\sim D,\mathbf{a}\sim\pi_\beta\left(\mathbf{a}|\mathbf{s}\right)}\left[Q\left(\mathbf{s},\mathbf{a}\right)\right]\right)\nonumber\\
&+&\frac{1}{2}\mathbb{E}_{\mathbf{s},\mathbf{a},\mathbf{s}^\prime\sim D}\left[\left(Q\left(\mathbf{s},\mathbf{a}\right)-\mathcal{B}^*Q\left(\mathbf{s},\mathbf{a}\right)\right)^2\right],
\end{eqnarray}

where $\alpha_\text{CQL}$ controls the conservative term.

\begin{figure}[h]
	\centering
	\includegraphics[width=\linewidth]{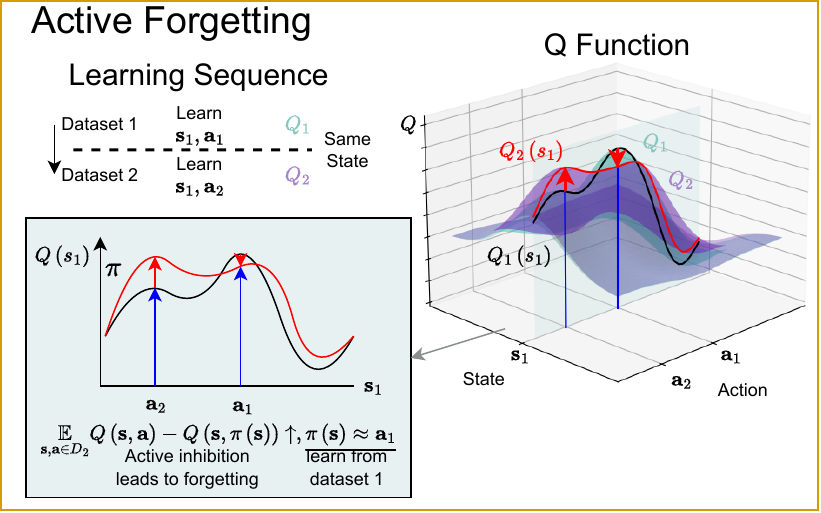}
	\caption{The diagram of active forgetting. In this picture, the network learns two datasets sequentially. Each of them has data point $\left(\mathbf{s}_1,\mathbf{a}_1\right)$ and $\left(\mathbf{s}_1,\mathbf{a}_2\right)$ respectively. Learning a worse action after a better action sequentially will result in forgetting directly. This kind of forgetting is not affected by the distribution shift.}
	\label{fig:STCORL_ActiveForgetting1}
\end{figure}

Conservative learning process well in an offline RL setting. However, when we introduce it into the DICORL setting, it will face a new kind of problem, active learning. In DICORL, when facing a dataset, the agent will actively suppress the Q values of the data deemed OOD relative to the dataset currently under study. As shown in Fig.~\ref{fig:STCORL_ActiveForgetting1}, after learning $\left(\mathbf{s}_1,\mathbf{a}_1\right)$ in the first dataset, the network learns another data point with the same (or very similar) state but with a different action $\mathbf{a}_2$ in the following dataset. Because the network cannot distinguish whether $\left(\mathbf{s}_1,\mathbf{a}_1\right)$ is real OOD data (out of all learnt datasets), or has been learnt in a previous dataset, it will restrain the Q value of the learnt data $\left(\mathbf{s}_1,\mathbf{a}_1\right)$, which will lead the policy network to avoid choosing $\mathbf{a}_1$ at $\mathbf{s}_1$, and will result in forgetting. We call this type of forgetting "active forgetting" because the network actively forgets the learnt action. Because this problem is caused by conservative learning, we also named it "conservative forgetting". We define active forgetting as the phenomenon that the agent forgets a learnt knowledge because it actively penalise the knowledge, but not the passive distribution shift. It is easy to see that only the DICORL will face the active forgetting problem, because firstly only the single task learning problem needs to learn knowledge of the same state repeatedly and secondly only the offline RL problem needs to select the best action from the input dataset. Notice that although algorithms such as implicit Q learning (IQL) \cite{kostrikovOfflineReinforcementLearning2021a} do not explicitly suppress the Q value, they still lead to active forgetting. IQL learns the Q network and the value network V through expectile regression and the policy network through advantage weighted regression (AWR) to avoid overestimation:

\begin{eqnarray} 
L_V&=&\mathbb{E}_{\left(\mathbf{s},\mathbf{a}\right)\sim D}\left[L_2^\tau\left(Q\left(\mathbf{s},\mathbf{a}\right)-V\left(\mathbf{s}\right)\right)\right],\\
L_Q&=&\mathbb{E}_{\left(\mathbf{s},\mathbf{a},\mathbf{s}^\prime\right)\sim D}\left[\left(r\left(\mathbf{s},\mathbf{a}\right)+\gamma V\left(\mathbf{s}^\prime\right)-Q\left(\mathbf{s},\mathbf{a}\right)\right)\right],
\end{eqnarray}
where $L_2^\tau(u)=\left|\tau-\mathbf{1}\left(u<0\right)\right|u^2$ denotes the expectile loss and $\tau$ the expectile threshold. According to \cite{kostrikovOfflineReinforcementLearning2021a}, the expectile threshold $\tau$ ranges from 0.7 to 0.9. Learning the low-quality dataset will cause active forgetting by overriding the knowledge learned before, even though the speed is slow because of the expectile threshold (about 0.3 to 0.1 times based on the $\tau$).

Active forgetting in the policy network may concurrently arise with that in the critic network. The policy network only learns data from the newest datasets. For IQL, the loss of the policy is:

\begin{equation} 
L_\pi=\mathbb{E}_{\left(\mathbf{s},\mathbf{a}\right)\sim D}\left[\exp\left(\alpha\left(Q\left(\mathbf{s},\mathbf{a}\right)-V\left(\mathbf{s}\right)\right)\right)\log\pi\left(\mathbf{a}|\mathbf{s}\right)\right],
\end{equation}

where $\alpha$ denotes the advantage weighting coefficient.

Although AWR used in IQL can alleviate learning of suboptimal actions, as only data from the present dataset undergo learning, inferior datasets still overwrite acquired knowledge from previously learnt datasets.

\begin{figure}[htbp]
	\centering
	\includegraphics[width=\linewidth]{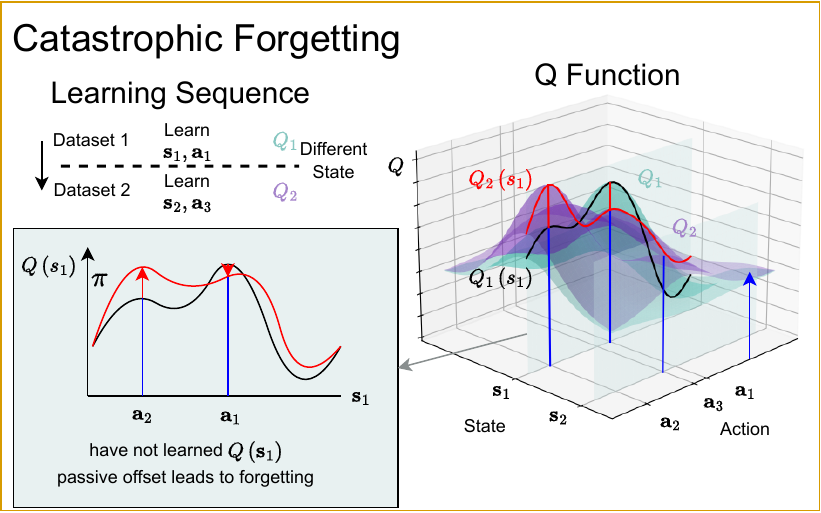}
	\caption{The diagram of the catastrophic forgetting. In this picture, the network needs to learn two datasets sequentially. Each of them has data point $\left(\mathbf{s}_1,\mathbf{a}_1\right)$ and $\left(\mathbf{s}_2,\mathbf{a}_2\right)$ respectively. We can see that even though these two points have different states, learning the following one will also affect the action selected in the previous state because of the distribution shift.}
	\label{fig:STCORL_CatastrophicForgetting}
\end{figure}

We would like to emphasise the difference between active forgetting and catastrophic forgetting in DICORL. Although active forgetting in DICORL is focused in this work, catastrophic forgetting also exists in DICORL. As shown in Fig.~\ref{fig:STCORL_CatastrophicForgetting}, catastrophic forgetting denotes after learning $\left(\mathbf{s}_1,\mathbf{a}_1\right)$, the network learns another data of \textbf{different} state $\left(\mathbf{s}_2,\mathbf{a}_2\right)$ in the following dataset, but affects the Q value at $\left(\mathbf{s}_1,\mathbf{a}_1\right)$. In comparison, active forgetting denotes the \textbf{superior} action (action with high Q value) on the \textbf{same} state space input getting overridden by the \textbf{inferior} action (action with low Q value) from learning a suboptimal new data actively, actively and catastrophic forgetting does not constitute independent issues but rather mutually reinforce each other.

Notice that only covering superior action with inferior action is a kind of forgetting, the opposite is the right learning process. In addition, after introducing the continual learning method, the network may refuse to learn the superior action after learning inferior action because of excessive stability. We called this kind of refuse "active rejection" and consider it as a kind of active forgetting.

\subsection{Ensemble Implicit Q Learning}
Most offline RL algorithms lead to active forgetting. Among the commonly used offline RL algorithms, as far as we know, SAC-N and EDAC \cite{anUncertaintyBasedOfflineReinforcement2021} do not suffer active forgetting. Considering that conservative learning aims to prevent overestimated initial Q values for OOD states, which causes overestimation, SAC-N introduces ensemble Q learning with multiple critic networks, using the minimum output among networks as the Q value for each state. By avoiding actively diminishing Q values, active forgetting is avoided at the same time.

However, ensemble methods (including follow-up work like MSG \cite{ghasemipourWhyPessimisticEstimating2022} and LB-SAC \cite{nikulinQEnsembleOfflineRL2023}) cannot achieve the best DICORL performance. This is primarily because IQL does not utilise OOD data during learning, so catastrophic forgetting mutually arising in actor and critic networks exerts limited influence, avoiding triggering severe active forgetting. Furthermore, the policy network training approach adopted in SAC-N is more unstable than the AWR used in IQL\cite{hansen-estruchIDQLImplicitQLearning2023}.
\begin{figure}[htbp]
	\centering
	\includegraphics[width=\linewidth]{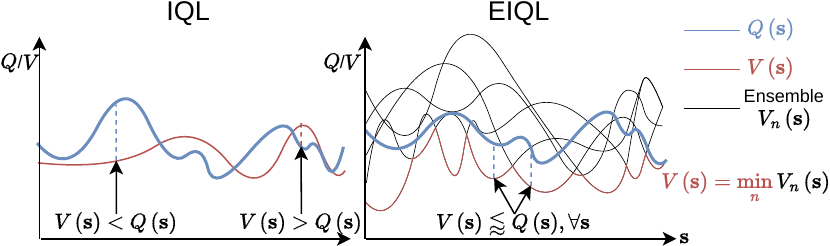}
	\caption{The diagram of the EIQL. By using ensemble value networks, EIQL keeps the initialized value network lower than the Q network at any state, so that the EIQL can use a very small $\tau$ to avoid active forgetting.}
	\label{fig:STCORL_EIQL}
\end{figure}

Therefore, we combine SAC-N and IQL into a new ensemble implicit Q learning method (EIQL). As shown in Fig.~\ref{fig:STCORL_EIQL}, EIQL initialises multiple value functions, assigning a sufficiently low value to each state. It then sustains the highest learned values through advantage weighted regression:
\begin{eqnarray}
L_V&=&\mathop{\mathbb{E}}\limits_{\left(\mathbf{s},\mathbf{a}\right)\sim D}\left[L_2^\tau\left(Q\left(\mathbf{s},\mathbf{a}\right)-\mathop{\mathbb{E}}\limits_jV^j\left(\mathbf{s}\right)\right)\right],\label{equ:STCORL_V}\\
L_Q&=&\mathop{\mathbb{E}}\limits_{\left(\mathbf{s},\mathbf{a},\mathbf{s}^\prime\right)\sim D}\left[\left(r+\gamma\min\limits_jV^j\left(\mathbf{s}^\prime\right)-Q\left(\mathbf{s},\mathbf{a}\right)\right)\right],\label{equ:STCORL_Q}
\end{eqnarray}
where $V^j,j=1,\dots,M$ denotes the $j^\text{th}$ value function. Hence, only the minimum value function is leveraged in EIQL. The expectation in Eq.~\ref{equ:STCORL_V} is introduced for robustness and generalisation, which do not affect the use of the minimum value function.
Through this approach, sufficiently low initialisation of the value functions for OOD data enables a higher expectile threshold $\tau$ than EIQL, which can alleviate active forgetting as shown above. In our experiment, $\tau$ is set to $0.99$ for EREIQL. In contrast, for EIQL, $tau$ is from $0.7$ to $0.9$.

On the other hand, to address active forgetting in the policy network, EIQL introduces a cloned policy network, merging its outputs with the new dataset for learning. Assuming the network finished learning dataset $n-1$, started on dataset $n$, for state $\mathbf{s}$, either the policy network $\pi_n$ selects the best action in the new data or maintains the previous $\pi_{n-1}$:
\begin{eqnarray}
L_{\pi_n}&=&\mathbb{E}_{\left(\mathbf{s},\mathbf{a}\right)\sim D_n}\left[\exp\left(\alpha\Delta\left(\mathbf{a},\mathbf{s}\right)\right)p\left(\mathbf{a}|\mathbf{s}\right)\right]\nonumber\\
&+&\mathbb{E}_{\left(\mathbf{s}\right)\sim D_n}\left[\exp\left(\alpha\Delta\left(\mathbf{a}^\prime,\mathbf{s}\right)\right)\log\pi_n\left(\mathbf{a}^\prime|\mathbf{s}\right)\right],
\end{eqnarray}\label{equ:STCORL_pi}
where $\mathbf{a}^\prime=\pi_{n-1}$ denotes the action from the cloned network, $\Delta\left(\mathbf{a},\mathbf{s}\right)=Q\left(\mathbf{s},\mathbf{a}\right)-\min_jV^j\left(\mathbf{s}\right)$ is different of the Q function and the value function, $\alpha$ is the coefficient, and $p\left(\mathbf{a}|\mathbf{s}\right)=\log\pi_n\left(\mathbf{a}|\mathbf{s}\right)$ is the logistic probability of the policy. In this way, the cloning of the policy network is controlled by the Q network, so that the inferior action in the replay buffer will not affect the performance.

Finally, existing continual learning algorithms still need to be introduced into the critic, value and policy networks of EIQL to mitigate catastrophic forgetting. In particular, according to \cite{wolczykContinualWorldRobotic2021}, continual learning algorithms prove inadequate for critic networks in task-incremental continual reinforcement learning, because only the policy network will be leveraged after learning one task. However, continual learning of critic networks is imperative for DICORL, since during future dataset learning the assistance of the critic network is needed.

We choose experience replay (ER) \cite{rolnickExperienceReplayContinual2019} as the continual learning method according to \cite{wolczykDisentanglingTransferContinual2022}. Subsequent experiments will demonstrate that ER is the most effective continual learning approach for DICORL. For data selection, we adopt the average Q value per trajectory as the rank method like \cite{hayesSelectiveReplayEnhances2021}. Although more advanced strategies such as \cite{gaiOfflineExperienceReplay2023} may perform better, selecting these functions is beyond the main focus of our work. By integrating EIQL and ER, our algorithm is termed experience-replay-based ensemble implicit Q learning (EREIQL).


\section{Experiments and Results}
We design the network following \cite{10.5555/3586589.3586904}. Please refer to Appendix A for detailed information.

\paragraph{Baseline Algorithms}
We contrast two algorithm groups as the baselines to assess the performance of existing algorithms: Continual learning algorithm baselines, including BC\cite{rolnickExperienceReplayContinual2019}, Averaged gradient episodic memory (AGEM)\cite{chaudhryEfficientLifelongLearning2019}, Elastic weight consolidation (EWC)\cite{kirkpatrickOvercomingCatastrophicForgetting2017}, Synaptic intelligence (SI)\cite{zenkeContinualLearningSynaptic2017}, and Riemannian walk (R-Walk)\cite{chaudhryRiemannianWalkIncremental2018}; and Offline RL algorithm baselines, including TD3+BC \cite{fujimotoMinimalistApproachOffline2021}, IQL \cite{kostrikovOfflineReinforcementLearning2021a}, SAC-N \cite{anUncertaintyBasedOfflineReinforcement2021} and EDAC \cite{anUncertaintyBasedOfflineReinforcement2021}. Please refer to Appendix B for more detailed information.

\paragraph{Offline Datasets}
We use four offline datasets from \cite{fuD4RLDatasetsDeep2020}: Hopper, HalfCheetah, Walker2D, and Ant. For each task, the network needs to learn a total of nine datasets, in the order Random1-Random2-Random3-Medium1-Medium2-Medium3-Random4-Random5-Random6. The number here means the index of a individual dataset, Random1 means the first random dataset, Medium1 means the first medium dataset, etc. The purpose of this training is to test three abilities of the algorithm, which are the ability to improve performance when learning a better dataset after learning a worse dataset (plasticity), the ability to maintain performance when learning a worse dataset after learning a better dataset (stability), and the ability to improve performance when learning different datasets with the same quality.
\paragraph{Metrics}
Following \cite{KernelCL}, we adopt the average performance (PER) and the backward transfer (BWT) as evaluation metrics,
\begin{equation}
\text{PER}=\frac{1}{N}\sum\limits_{n=1}^Na_{N,n},\ \text{BWT}=\frac{1}{N-1}\sum\limits_{n=1}^{N-1}a_{n,n}-a_{N,n},
\end{equation}
where $a_{i,j}$ means the final cumulative rewards of dataset $j$ after learning dataset $i$. For PER, higher is better; for BWT, lower is better.
These two metrics show the ability of a continual learning algorithm to learn new tasks while alleviating the problem of forgetting.

Also, we propose that it is necessary to report the final performance after learning all the datasets as another reference in DICORL:
\begin{equation}
\text{LST}=a_{N,N}.
\end{equation}
In the following, we use LST to represent this metric. Same as PER, for LST, higher is better.

\subsection{Reinforcement Learning Algorithm in DICORL}


\begin{table*}[htbp]
	\centering
	\caption{Results of different algorithms in DICORL setting. In which LST is the last performance, PER is the mean performance, and BWT is the mean backward translation. For LST and PER, higher is better; for BWT, lower is better.}
	\begin{tabular}{ccccccc}
		\hline
		\multicolumn{2}{c}{Benchmark}&TD3+BC&SACN&EDAC&IQL&EIQL\\
		\hline
		\multirow{3}{*}{HalfCheetah}&LST$\uparrow$&1704.80&-474.56&-596.00&1422.21&\textbf{3996.05}\\
		&PER$\uparrow$&\textbf{3892.53}&-513.12&-386.90&2463.16&3385.72\\
		&BWT$\downarrow$&2625.30&224.95&295.47&3158.92&2625.30\\
		\hline
		\multirow{3}{*}{Hopper}&LST$\uparrow$&39.88&248.17&36.26&327.58&\textbf{1852.40}\\
		&PER$\uparrow$&963.80&108.56&38.75&618.85&\textbf{1070.68}\\
		&BWT$\downarrow$&2042.40&292.70&77.45&650.26&1573.92\\
		\hline
		\multirow{3}{*}{Walker2D}&LST$\uparrow$&-8.90&-2.20&-14.60&2692.34&\textbf{3230.55}\\
		&PER$\uparrow$&56.34&8.18&79.28&1578.40&\textbf{1716.75}\\
		&BWT$\downarrow$&165.50&132.45&613.93&1851.49&1760.88\\
		\hline
		\multirow{3}{*}{Ant}&LST$\uparrow$&1429.68&-1545.45&-1805.39&1493.43&\textbf{2029.74}\\
		&PER$\uparrow$&1083.21&-2035.68&-2079.81&1160.45&\textbf{1475.53}\\
		&BWT$\downarrow$&\textbf{1257.76}&1962.42&758.42&707.21&813.53\\
		\hline
	\end{tabular}
	\label{tab:STCORL_rl_results}
\end{table*}

Tab.\ref{tab:STCORL_rl_results} illustrates the DICORL performance of different RL algorithms with ER. As shown, the SAC-N and EDAC algorithms cannot maintain performance, which shows that their Q optimisation policy is unstable and unreliable in sequential learning. Moreover, neither TD3+BC nor IQL achieves adequate DICORL results. Compared to the inferior plasticity of IQL, which hinders learning to higher levels, TD3+BC demonstrates far worse stability, with performance sharply decreasing when exposed to random tasks, showing that the behaviour cloning method harm the final result deeply. In general, these existing offline RL algorithms fail at DICORL, because all of these algorithms are not designed for DICORL setting. In contrast, the proposed EREIQL achieves the best performance, successfully tackling DICORL. We would like to emphasise that the low BWT of other algorithms is not caused by the ability to relieve catastrophic forgetting or active forgetting, but by the low ability to learn all of the tasks and little space to reduce, which shows that the BWT is not a fitting metric for those tasks in which some baseline may totally fail.

\subsection{Continual Learning Algorithm in DICORL}  

\begin{figure}[htbp]
	\includegraphics[width=0.85\linewidth]{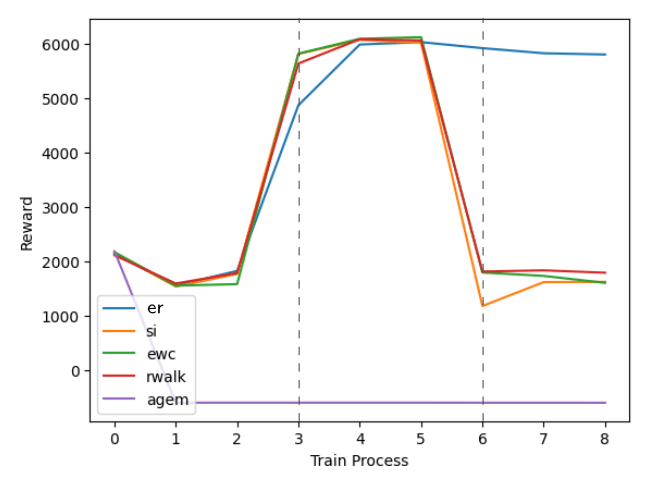}
	\caption{Performance of HalfCheetah across continual learning algorithms. The network learns a task for 500 epochs and turns to the next. The two dotted lines from left to right represent the switch from random to medium and from medium to random, respectively. Higher is better.}
	\label{fig:STCORL_cl}
\end{figure}

Fig.~\ref{fig:STCORL_cl} presents different DICORL performance of continual learning algorithms. As shown, apart from experience replay, other rehearsal-based and regularisation-based algorithms cannot achieve adequate DICORL performance. This decrease is in agreement with the findings of \cite{wolczykContinualWorldRobotic2021} that existing continual learning algorithms perform poorly on critic networks, which is necessary in DICORL setting. In TICORL, continually learning the critic networks is not necessary, so that they can avoid this hard challenge easily. But this kind of avoiding is not allowed in DICORL, so that the performance decrease quickly when learning on low quality dataset. Notice that because we are only training on a single task, dynamic-architecture-based continual learning methods are not applicable.

\subsection{Impact of Storage Space} 

\begin{figure}[htbp]
	\includegraphics[width=0.85\linewidth]{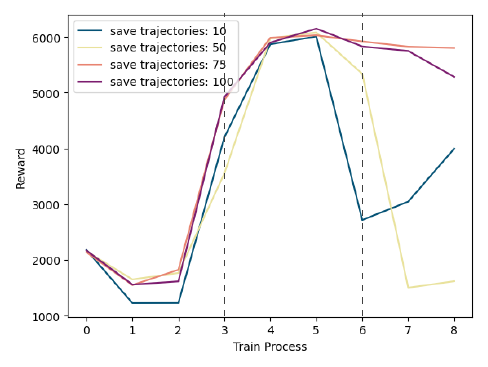}
	\caption{Performance of HalfCheetah across the different sizes of the replay buffer. The network learns a task for 500 epochs and turns to the next.  The two dotted lines from left to right represent the switch from random to medium and from medium to random, respectively. Higher is better.}
	\label{fig:STCORL_100_save}
\end{figure}

Notice that we select a huge replay buffer in our experiment. As shown in Fig.~\ref{fig:STCORL_100_save}, compared to the task-incremental continual RL in \cite{gaiOfflineExperienceReplay2023,wolczykContinualWorldRobotic2021}, DICORL poses substantially greater difficulty. To maintain learnt performance, algorithms require at least 75 trajectories versus just one in \cite{gaiOfflineExperienceReplay2023} because of the needs of continually critic networks. In other words, DICORL is a harder task than TICORL and waiting for further research. This can also analogy with the relationship of task-incremental learning and class-incremental learning in category tasks.

\begin{figure}[htbp]
	\includegraphics[width=0.85\linewidth]{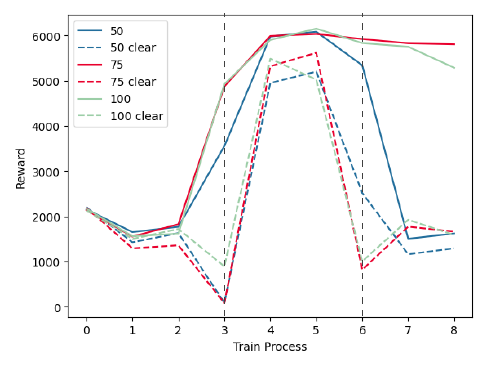}
	\caption{Performance of EREIQL in Half-Cheetah without continual learning. The network learns a task for 500 epochs and turns to the next. Solid lines indicate continual learning and dashed lines denote re-initialization before learning each dataset while retaining the replay buffer. The two dotted lines from left to right represent the switch from random to medium and from medium to random, respectively. Higher is better.}
	\label{fig:STCORL_single_learning}
\end{figure}

Notice that 75 trajectories is not a trifling amount. A plausible hypothesis holds that EREIQL does not resolve DICORL challenges but rather owes its performance solely to the considerable replay buffer size, which essentially constitutes a mixed dataset where EREIQL happens to boast superior learning capabilities over other algorithms. To eliminate this possibility, Fig.~\ref{fig:STCORL_single_learning} additionally provides results comparing continual learning and learning from scratch, with solid lines indicating continual learning and dashed lines denoting reinitialisation before learning each dataset while retaining the replay buffer. We can see that EREIQL acquires and maintains the best performance through continual single-task learning instead of merely possessing stronger mixed-dataset learning abilities. 

\subsection{Impact of Network Ensemble Size} 

\begin{figure}[htbp]
	\includegraphics[width=0.85\linewidth]{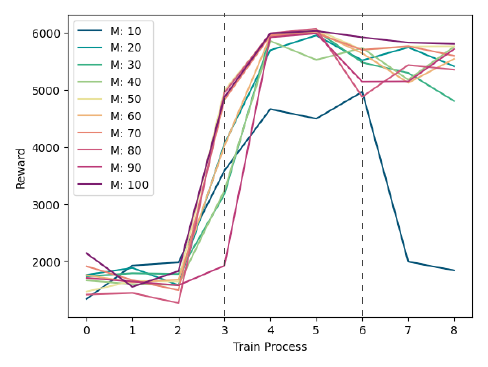}
	\caption{Performance of EREIQL in Half-Cheetah across different ensemble numbers. The network learns a task for 500 epochs and turns to the next. The two dotted lines from left to right represent the switch from random to medium and from medium to random, respectively. Higher is better.}
	\label{fig:STCORL_ensemble_num}
\end{figure}

Fig.\ref{fig:STCORL_ensemble_num} shows the impact of the number of value networks in EREIQL with the storage space fixed at 75 trajectories. EREIQL achieves strong performance from 30 to 100 networks. Compared to SAC-N, EREIQL achieves higher performance with fewer networks.  

\begin{figure}[htbp]
	\includegraphics[width=0.85\linewidth]{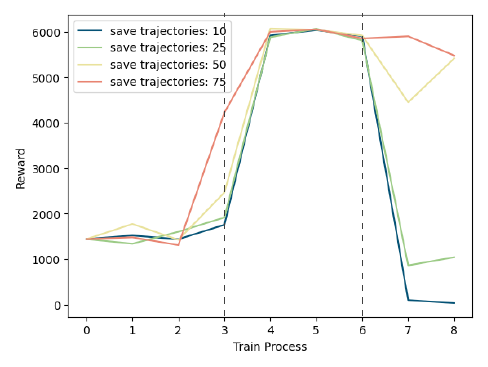}
	\caption{Performance of HalfCheetah across the different sizes of the replay buffer with 1000 ensemble value networks. The network learns a task for 500 epochs and turns to the next. The two dotted lines from left to right represent the switch from random to medium and from medium to random, respectively. Higher is better.}
	\label{fig:STCORL_1000_save}
\end{figure}

This discovery motivates the conjecture that more networks could reduce the demand for replay buffers. To validate this, Fig.\ref{fig:STCORL_1000_save} presents the size of the replay buffer effects on network performance with 1000 value networks, which shows that more networks reduce the reliance of EREIQL on the replay buffer.  

\subsection{Impact of Expectile Threshold $\tau$} 

\begin{figure}[htbp]
	\includegraphics[width=0.85\linewidth]{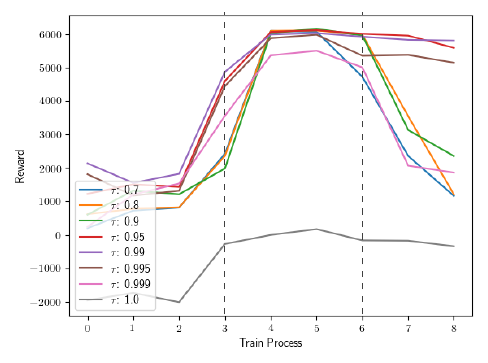}
	\caption{Performance of Half-Cheetah across different expectile regression threshold $\tau$. The network learns a task for 500 epochs and turns to the next.  The two dotted lines from left to right represent the switch from random to medium and from medium to random, respectively. Higher is better.}
	\label{fig:STCORL_expectile}
\end{figure}

As described, the primary advantage of EIQL over IQL lies in enabling higher $\tau$ values through ensemble learning to alleviate active forgetting. To validate this, Fig.~\ref{fig:STCORL_expectile} shows the impact of varying $\tau$. EIQL can set $\tau \geq 0.99$ without hindering implicit $Q$-learning, while IQL is limited to around 0.9. Larger $\tau$ values allow IQL greater certainty in acquired value functions, avoiding active forgetting. The capacity to utilize larger $\tau$ constitutes the core factor behind the efficacy of EREIQL in resolving DICORL.  

\section{Conclusion and Future Work}

In this work, we propose a new continual learning setting, DICORL. Different from the traditional TICORL, in DICORL, the agent should learn from a sequence of offline datasets of a single offline RL task and reach the best performance. We propose that in DICORL, the agent must address a new problem, active forgetting, which is caused by the different data quality of different offline datasets and the most important technology used by offline RL, conservative learning or behaviour cloning. Active forgetting is a fresh new problem, which is beyond the capability of traditional stability-plasticity balance that continual learning algorithms usually focus on. To address this issue, we propose a new offline RL approach based on ensemble learning, EREIQL, which improves the plasticity of new data through ensemble learning to allow strong stability on old data and high plasticity on new data. Our experiments firstly prove that DICORL is more challenging than common TICORL, and prevailing continual learning algorithms fail in more general continual learning settings. Then, the effectiveness of EREIQL is validated. However, the current EREIQL still has high demands on time and space complexity, and cannot suit more complex offline RL tasks. The next step will be further research on feasible and efficient continual and RL algorithms applied in DICORL settings.

\bibliography{main}
\bibliographystyle{icml2025}

\end{document}